# Quantifying Climate Change Impacts on Renewable Energy Generation: A Super-Resolution Recurrent Diffusion Model

Xiaochong Dong, *Member*, *IEEE*, Jun Dan, Yingyun Sun, *Member*, *IEEE*, Yang Liu, Xuemin Zhang, *Member*, *IEEE*, Shengwei Mei, *Fellow*, *IEEE*, *Fellow*, *CSEE*

*Abstract*—**Driven by global climate change and the ongoing energy transition, the coupling between power supply capabilities and meteorological factors has become increasingly significant. In the long term, accurately quantifying renewable energy generation under the influence of climate change is crucial for the development of sustainable power systems. However, due to interdisciplinary differences in data requirements, climate data often lacks the hourly resolution and fail to capture short-term uncertainty of renewable energy resources. To address this limitation, a super-resolution recurrent diffusion model (SRDM) has been developed to enhance the temporal resolution of climate data and model the short-term uncertainty. The SRDM integrates a pre-trained decoder and a denoising network within a recurrent coupling mechanism to generate long-term, high-resolution climate data. The high-resolution climate data is then converted into power generation using the mechanism model, enabling the simulation of wind and photovoltaic power generation on future long-term scales. Case studies were conducted in the Ejina region of Inner Mongolia, China, using coupled model intercomparison project (CMIP6) data to simulation three representative climate pathways. The results demonstrate that SRDM outperforms existing generative models in generating super-resolution climate data. Furthermore, the research highlights the estimation biases introduced when low-resolution climate data is used for power conversion.**

## I. INTRODUCTION

TRADITIONAL power systems have historically relied on fossil fuel-based electricity generation, which is central to modern socio-economic operations but also remains the largest source of global carbon dioxide emissions [1]. In the context of the energy transition, power systems must ensure the provision of safe, economical, and low-carbon electricity. In response, governments worldwide are implementing policies to promote the development of renewable energy, leading to a rapid increase in the installed capacities of wind and photovoltaic (PV) power. This shift is transforming power systems from weakly coupled meteorological systems to those that are strongly coupled [2]. To meet the carbon emissions reduction goals for the 21st century, it is essential to accurately quantify the power generation of renewable energy sources on long-term scales. It can effectively analyze the supply and demand balance within power systems and provide a reliable foundation for system planning and decision-making.

Modeling the long-term power generation capability of renewable energy primarily relies on three types of data: historical measurement data, reanalysis data, and climate data. The following sections discuss the application methods, advantages, and limitations of each data type.

Historical measurement data are typically obtained from system operators or research institutions, such as the Belgian transmission system operator Elia [3] or the National Renewable Energy Laboratory in the United States [4]. This data provides the most accurate representation of fluctuations and trends in renewable energy generation. When combined with meteorological measurements, it allows for the analysis of meteorology to power conversion and is often characterized by high temporal resolution, with sampling intervals ranging from 5 to 15 minutes in most regions. However, the accumulation of observational data requires extended periods. The available data for renewable power generation rarely exceeds 20 years, and many newly built wind and solar farms have less than five years of recorded measurements. Furthermore, maintenance and calibration issues with monitoring equipment can lead to declining measurement accuracy or long-term data loss [5]. In certain undeveloped or unmonitored areas, corresponding measurement data may be entirely unavailable. While adjacent areas with similar features can be used for data transfer, this approach may be unreliable due to significant domain shifts in certain regions [6]. Additionally, advancements in energy technology have reduced the availability of historical data. New equipment, such as perovskite solar cells [7] or large-capacity wind turbines [8], exhibit power generation characteristics that differ significantly from those of earlier systems.

Reanalysis datasets, which integrate meteorological observations with numerical weather prediction models, offer global simulations of historical meteorological conditions. Typical dataset includes the European centre for medium-range weather forecasts (ECMWF) fifth-generation reanalysis (ERA5)

This work was supported in part by the National Key Research and Development Program of China under Grant 2022YFB2403000, in part by the Postdoctoral Fellowship Program and China Postdoctoral Science Foundation under Grant BX20250414 and in part by the China Postdoctoral Science Foundation under Grant 2025M770478.

X. Dong, X. Zhang *(corresponding author)* and S. Mei are with the State Key Laboratory of Power System Operation and Control, Department of Electrical Engineering, Tsinghua University, Beijing, 100084, China (e-mail: dream_dxc@mail.tsinghua.edu.cn; zhangxuemin@mail.tsinghua.edu.cn; meishengwei@mail.tsinghua.edu.cn)

J. Dan is with the College of Information Science & Electronic Engineering, Zhejiang University, Hangzhou 310058, China (e-mail: danjun@zju.edu.cn)

Y. Sun is with the State Key Laboratory of Alternate Electrical Power System with Renewable Energy Sources, North China Electric Power University, Beijing 102206, China (e-mail: sunyy@ncepu.edu.cn)

Y. Liu is with the Department of Engineering, King's College London, London WC2R 2LS, UK (e-mail: yang.15.liu@kcl.ac.uk)

[9] and national aeronautics and space administration (NASA) Modern-era retrospective analysis for research and applications (MERRA-2) [10]. These datasets cover periods of up to 80 years and enable the simulation of wind and PV power generation at specific locations by transforming meteorological data into power generation data [11]. Wind power is typically estimated by adjusting wind speed to turbine hub height using a wind profile model, followed by conversion to theoretical power based on the wind turbine power curve [12]. PV power generation is assessed based on solar irradiance received by the module and accounting for efficiency losses due to temperature variations [13]. However, in real-world applications, numerous additional factors influence wind and solar power output, such as PV power fluctuations caused by cloud cover and efficiency losses in wind farms due to turbine wake effects. With a temporal resolution of up to one hour, reanalysis data are well-suited for power system studies involving supply-demand balance, and long-term planning scenarios [14–15].

However, both historical measurement and reanalysis data describe historical conditions without accounting for the impact of future climate change on renewable energy resources. In power system planning and operation, analyses based on these datasets often rely on the assumption of climatic stationarity. On long-term timescales, however, global climate change leads to a redistribution of wind and solar resources, altering the power generation capability of renewable energy across different regions [16-17]. Quantifying the long-term renewable energy generation under climate change is critical. It not only affects the generation of existing renewable resources but also informs planning decisions for future generation resources during the energy transition [18-19].

Climate data are derived from global climate model (GCM), which simulate long-term trends in wind speed, solar radiation, and other meteorological features under different radiative forcing pathways. Similar to reanalysis data, when combined with mechanism model of wind turbines and PV modules, climate data can be used to estimate future renewable energy generation [20]. Studies indicate that global warming is associated with a general decline in wind energy resources, particularly in mid-latitude regions of the northern hemisphere, while solar energy resources remain relatively stable or exhibit a slight upward trend [21-22]. However, the use of climate data for renewable energy modeling faces several challenges: (1) **Temporal Resolution:** The primary focus of the climate science community is on analyzing long-term trends over large spatial scales. GCM typically have temporal resolutions of daily or monthly, which are insufficient for capturing the short-term variability of renewable energy. Power systems require higher temporal resolution (typically at hourly or sub-hourly intervals) for supply-demand balance analyses and to verify the short-term ramping capabilities of thermal power units and other flexible resources [23-24]. (2) **Uncertainty Representation:** Climate data provide deterministic simulations for specific radiative forcing pathways, failing to capture the short-term stochastic feature of renewable energy generation. As renewable energy penetration increases, short-term uncertainties exert a significant influence on the supply-demand balance in future power systems [25]. While techniques such as regional climate model (RCM) and ensemble forecasting can address these issues, they typically demand substantial computational resources, which limits their practicality for broader energy research applications. The divergence in data requirements between climate and energy sciences and the computational complexity of numerical solutions, presents significant challenges to efficiently generating detailed climate data suited for energy research, thereby hindering effective interdisciplinary integration [26].

To address the aforementioned challenges, this paper proposes a novel methodology for quantifying the power generation of renewable energy while accounting for the effects of climate change. The main contributions of this study are as follows:

(1) The study proposes a super-resolution recurrent diffusion model (SRDM) to enhance the temporal resolution of climate data to an hourly level, meeting the computational requirements for supply and demand balance analyses in power systems. The proposed recurrent diffusion model ensures continuity in time-series super-resolution and non-parametrically generates long-term renewable energy scenarios. This approach not only aligns with long-term climate change trends but also effectively captures the stochastic nature of short-term renewable energy generation.

(2) By integrating the mechanism model, the proposed approach effectively quantifies the power generation of renewable energy sources in undeveloped regions. Case studies are conducted in the Ejina region of Inner Mongolia, China, to analyze long-term trends in power generation under various climate change pathways. The study shows the potential for significant biases when using low-resolution, averaged climate data for simulations, emphasizing the critical need for high-resolution climate data modeling.

The rest of this manuscript is organized as follows. Section II outlines the problem formulation. Section III details the network structure of the SRDM. Section IV presents the case study methodology and numerical results. Finally, Section V concludes the manuscript and future research directions.

## II. Problem Formulation

Quantifying renewable energy generation under climate change impacts can be divided into three steps: global climate simulation, downscaled climate simulation, and power generation simulation, as shown in Fig. 1.

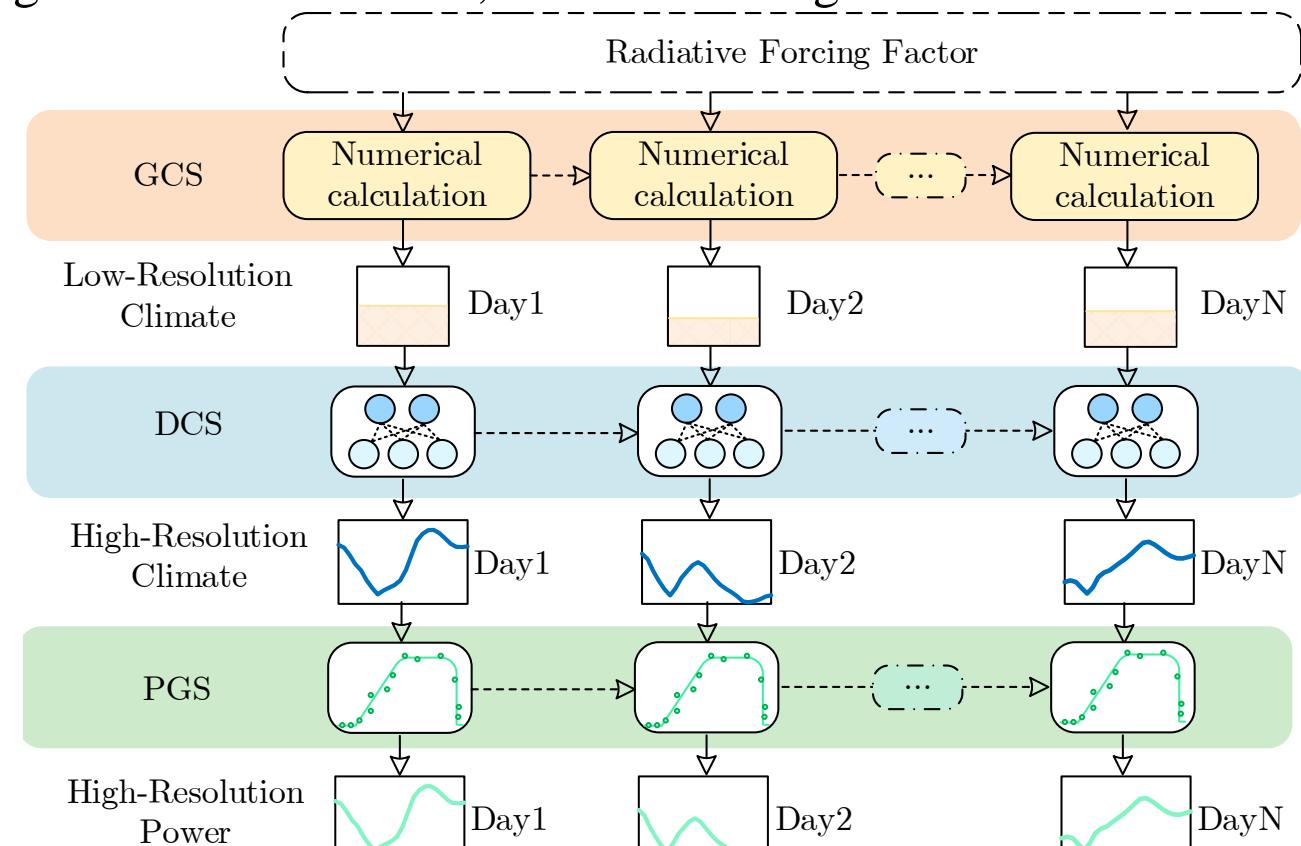

Fig. 1. Quantifying renewable energy generation.

To facilitate the subsequent formulation of equations, some fundamental expressions are defined as follows. Let $\overline{\boldsymbol{x}}_d$ denote low-resolution climate data with daily temporal resolution, and let $\boldsymbol{x}_d$ represent high-resolution climate data with hourly temporal resolution, where the subscripts $d$ indicate time index.

**(1) Global Climate Simulation (GCS)**

GCM typically use various radiative forcing factors as boundary conditions to simulate large-scale climate processes with spatial resolutions ranging from tens to hundreds of kilometers and temporal resolutions spanning days to months. The coupled model intercomparison project (CMIP6) provides extensive global climate datasets for research. Consider a simplified climate simulation process in which the climate data from the $(d\text{-}1)$ day serve as the initial condition for simulating the data of the $d$ day. This process can be expressed as:

$$\overline{\boldsymbol{x}}_d = f(\overline{\boldsymbol{x}}_{d-1}, c_{\text{ssp}}, \eta) \tag{1}$$

where $c_{\text{ssp}}$ is the boundary conditions of climate pathway, $\eta$ is the stochastic noise that describes the chaotic nature of GCS, and $f(\cdot)$ is the mapping function of GCS process.

**(2) Downscaled Climate Simulation (DCS)**

DCS is performed based on GCS results to achieve hourly temporal resolution, thereby capturing short-term temporal meteorological features more accurately. The combination of DCS and GCS not only reduces computational costs but also ensures consistency between high-resolution and low-resolution climate data. DCS uses the simulation results of GCS as the boundary condition and takes the $(d\text{-}1)$ day data as the initial condition. It simulates the downscaled climate processes using statistical or dynamic methods, expressed as:

$$\boldsymbol{x}_d = g(\boldsymbol{x}_{d-1}, \overline{\boldsymbol{x}}_d, \xi) \tag{2}$$

where is $\xi$ the stochastic noise that describes the chaotic nature of DCS, and $g(\cdot)$ is the mapping function of DCS process.

The downscaling process shares similarities with the super-resolution problem and can be viewed as an ill-posed problem. Essentially, it involves mapping high-dimensional probability distributions base on low-dimensional conditional values. By introducing generative models, the downscaling mapping process can be implemented using a conditional generative model, using $\overline{\boldsymbol{x}}_d$ and $\boldsymbol{x}_{d-1}$ as conditions.

**(3) Power Generation Simulation (PGS)**

The simulation of power generation employs mechanism models for wind turbine and PV module to describe the deterministic mapping relationship between meteorological variables and power generation. This process is expressed as:

$$\boldsymbol{p}_d = h(\boldsymbol{x}_d) \tag{3}$$

where $h(\cdot)$ is the mapping function of meteorological features and power generation.

## III. Super-resolution Recurrent Diffusion Model

### A. *Latent Diffusion Models*

Diffusion models are a class of deep generative models that restore data distribution $q(\boldsymbol{x})$ from a prior noise $q(\boldsymbol{z})$ by training a denoising neural network, where $\boldsymbol{x} \in \mathbb{R}^T$ and $\boldsymbol{z} \in \mathbb{R}^T$. In the vanilla diffusion model, the prior noise has the same dimensionality as the target data, and optimizing the probability transformation path in high-dimensional manifold space incurs significant computational costs. To address this challenge, the latent diffusion model (LDM) was introduced. LDM uses variational auto-encoder (VAE) to map high-dimensional samples $\boldsymbol{x} \in \mathbb{R}^T$ into a low-dimensional latent space $\boldsymbol{z} \in \mathbb{R}^{T'}$, where $T < T'$ [27]. The VAE encoder decouples correlations between dimensions, bringing samples closer to prior noise in the low-dimensional manifold space. Consequently, the diffusion model operates in the latent space, effectively reducing training complexity.

LDM invert a predefined diffusion process to recover the latent space distribution. The predefined diffusion process is formulated as a Markov process with stepwise noise diffusion, expressed as:

$$q(\boldsymbol{z}_n) = q(\boldsymbol{z}_0) \prod\nolimits_n q(\boldsymbol{z}_n \mid \boldsymbol{z}_{n-1}), 1 \le n \le N \tag{4}$$

where $\boldsymbol{z}_0$ is the latent space sample, $\boldsymbol{z}_n$ is the $n$-step diffusion latent space sample, $\boldsymbol{z}_N \sim \mathcal{N}(\boldsymbol{0}, \boldsymbol{1})$ is the prior noise, and $N$ is the nosing step.

The diffusion process can be described by a multivariate Gaussian distribution as:

$$q(\boldsymbol{z}_n \mid \boldsymbol{z}_{n-1}) \sim \mathcal{N}(\sqrt{1-\beta_n}\boldsymbol{z}_{n-1}, \beta_n \mathbf{I}), 1 \le n \le N \tag{5}$$

where $\beta_n$ is coefficients following a fixed schedule.

The corresponding denoising process is modeled as a neural network-based Markov process with stepwise denoising, expressed as:

$$p_\theta(\boldsymbol{z}_0) = q(\boldsymbol{z}_N) \prod\nolimits_n p_\theta(\boldsymbol{z}_{n-1} \mid \boldsymbol{z}_n) \tag{6}$$

The denoising process can also be described by a multivariate Gaussian distribution as:

$$p_\theta(\boldsymbol{z}_{n-1} \mid \boldsymbol{z}_n) \sim \mathcal{N}(\boldsymbol{\mu}_\theta(\boldsymbol{z}_n, n), \boldsymbol{\Sigma}_\theta(\boldsymbol{z}_n, n)) \tag{7}$$

where expectation $\boldsymbol{\mu}_\theta$ and covariance matrix $\boldsymbol{\Sigma}_\theta$ are predicted by the denoising neural network.

### B. *Structure of SRDM*

Based on the LDM, the SRDM model for realizing downscaled climate data is proposed. It consists of two components: a denoising network and a decoder network. These components collaboratively map low-resolution climate data to high-resolution ensemble climate data. The mapping process is recurrent, ensuring the continuity of long-term time-series data, as shown in Fig. 2.

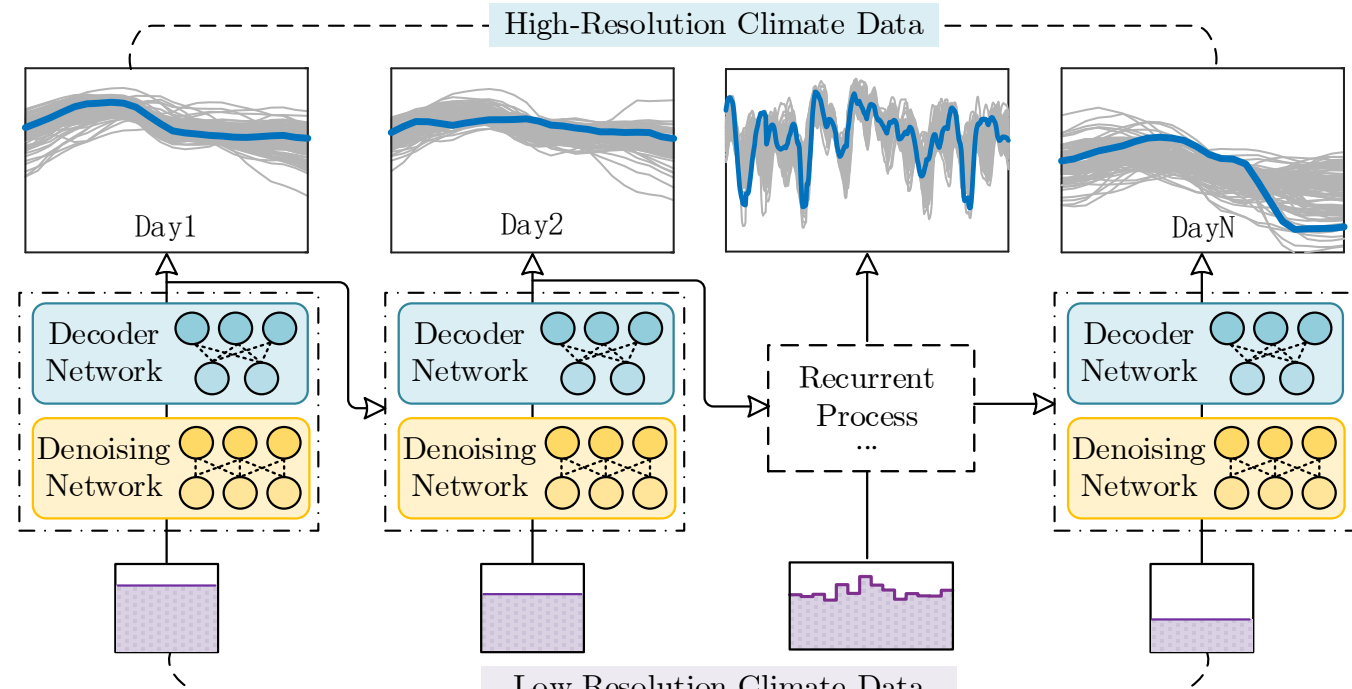

Fig. 2 Recurrent process of SRDM.

As shown in Fig. 2, the SRDM requires the high-resolution data from the previous day as an initial condition to ensure temporal continuity. Simultaneously, the low-resolution climate data serve as boundary conditions to maintain consistency between the high- and low-resolution climate data. Leveraging the stochastic nature of diffusion models, the generated datasets exhibit short-term uncertainty. The mapping process is expressed as:

$$\begin{cases} \boldsymbol{z}_d = \mathrm{DN}(\boldsymbol{x}_{d-1}, \overline{\boldsymbol{x}}_d, \xi) \\ \boldsymbol{x}_d = \mathrm{VD}(\boldsymbol{z}_d) \end{cases} \tag{8}$$

where $\mathrm{DN}(\cdot)$ is the mapping function of denoising network, and $\mathrm{VD}(\cdot)$ is the mapping function of decoder network.

### C. Pre-Train Decoder Network

The decoder network in SRDM is derived from a pre-trained VAE, which decouples and reduces the dimensionality of high-dimensional time series data. The pre-trained VAE structure is shown in Fig. 3.

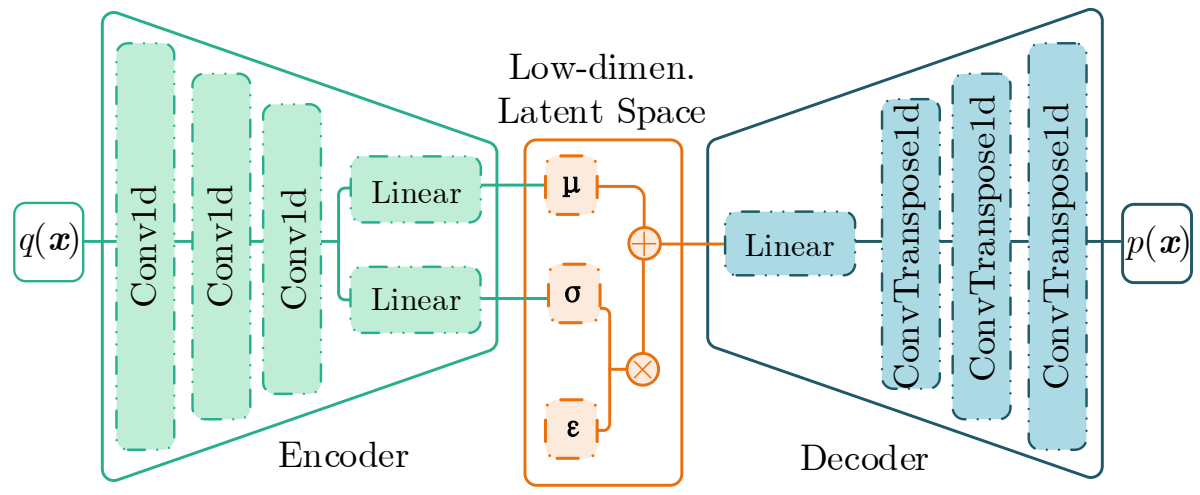

Fig.3 Structure of pre-trained VAE.

As shown in Fig.3, the VAE contains two networks: an encoder network and a decoder network. The encoder network uses stacked temporal convolutional layers for down-sample, obtaining the expectation and variance of the low-dimensional latent space via linear layers. The latent space sample distribution $p_\theta(\boldsymbol{z})$ as:

$$p_\theta(\boldsymbol{z}) \sim \mathcal{N}(\boldsymbol{\mu}_\theta(\boldsymbol{x}), \boldsymbol{\sigma}_\theta^2(\boldsymbol{x})) \tag{9}$$

where $\boldsymbol{\mu}_\theta$ and $\boldsymbol{\sigma}_\theta^2$ are the expectation and variance of the latent space. Monte Carlo sampling can be used to convert these values into latent space samples:

$$\boldsymbol{z} = \boldsymbol{\mu}_\theta(\boldsymbol{x}) + \boldsymbol{\sigma}_\theta(\boldsymbol{x}) \cdot \epsilon,\ \epsilon \sim \mathcal{N}(0,1) \tag{10}$$

The decoder network uses stacked $L$-layer temporal transposed convolutional layers for up-sampling to mapping the latent space sample distribution to a posteriori sample distribution. The $\boldsymbol{z}$ and $\boldsymbol{x}$ can form the prior joint distribution $q(\boldsymbol{x},\boldsymbol{z})$ and posteriori joint distribution $p_\theta(\boldsymbol{x},\boldsymbol{z})$, the training objective of the VAE can be regarded as minimizing the KL divergence between $q(\boldsymbol{x},\boldsymbol{z})$ and $p_\theta(\boldsymbol{x},\boldsymbol{z})$. The prior distribution of the latent space is the standard multivariate Gaussian distribution, so that the correlation between the dimensions is a decoupled state. The improved loss function of VAE can be transfer as:

$$\begin{aligned} L_{\mathrm{vae}} = \mathbb{E}_l \mid \mathrm{VE}^l(\boldsymbol{x}) - \mathrm{VD}^{L\text{-}l}(\mathrm{VE}(\boldsymbol{x})) \mid \\ + \lambda \cdot KL(p_\theta(\boldsymbol{z} \mid \boldsymbol{x}) \,||\, q(\boldsymbol{z})) \end{aligned} \tag{11}$$

where the $\mathrm{VE}(\cdot)$ is the mapping function of the encoder network, $l$ is the index of the hidden layer of the encoder and decoder, and $\lambda$ is the weight coefficient, which is used to balance the reconstruction error and KL divergence. The first term of the loss function is the reconstruction error losses, and the second term is the KL divergence between the posterior distribution $p_\theta(\boldsymbol{z} \mid \boldsymbol{y}) \sim \mathcal{N}(\boldsymbol{\mu}_\theta, \boldsymbol{\sigma}_\theta^2)$ of the latent space and the prior distribution $q(\boldsymbol{z}) \sim \mathcal{N}(\boldsymbol{0},\boldsymbol{1})$. The role of KL divergence term achieves a degree of temporal decoupling, thereby reducing the complexity of downstream tasks. In order to improve the accuracy of the VAE reconstruct samples, L1-base perceptual reconstruction loss is used to hidden layers [28].

The second term of KL divergence can be calculated by the following equation:

$$KL(p(\boldsymbol{z} \mid \boldsymbol{x}) \,||\, q(\boldsymbol{z}))] = \frac{1}{2}(\boldsymbol{\mu}_\theta^2 + \boldsymbol{\sigma}_\theta^2 - \ln \boldsymbol{\sigma}_\theta^2 - 1) \tag{12}$$

In the context of LDM, the primary role of the VAE is to perform dimensionality reduction rather than strictly enforcing that the latent space adheres to the prior distribution. Therefore, the weight coefficients $\lambda$ can be appropriately reduced to improve the ability of the pre-trained VAE to restore samples, which does not affect the mapping of the subsequent diffusion model.

### D. Denoising Network

The denoising network in SRDM is to realize the latent space mapping that satisfies the boundary condition and initial condition, and the structure of denoising network proposed is shown in Fig. 4.

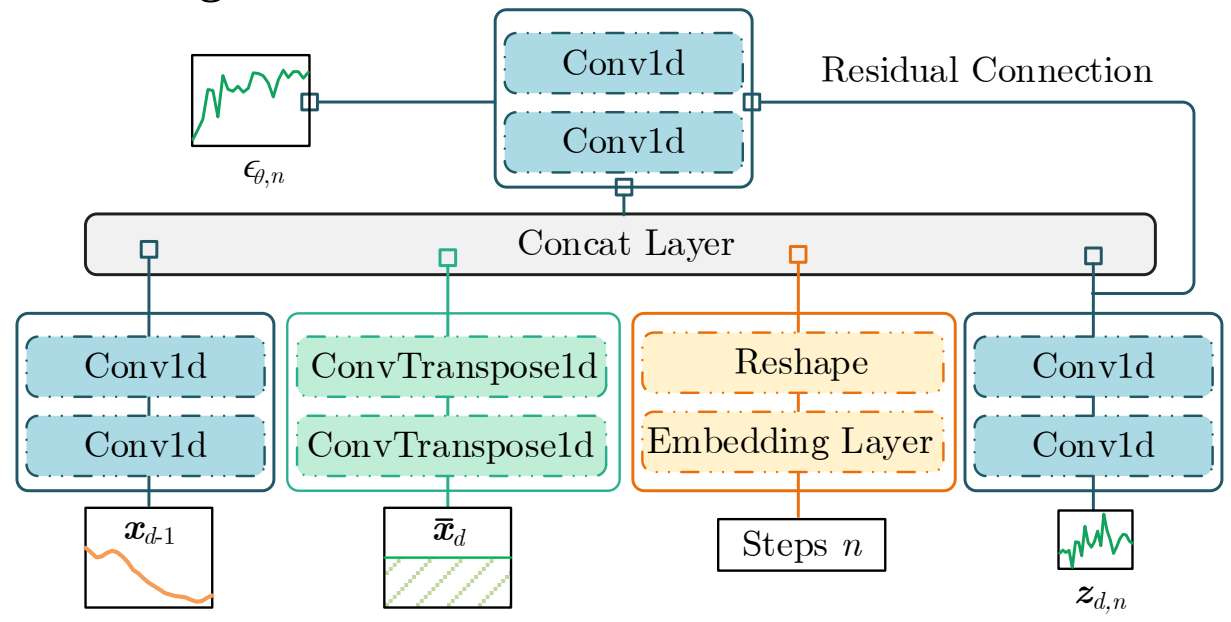

Fig.4 Structure of denoising network.

The denoising network contains several embedding modules, which are required to embed the $(d\text{-}1)$ day high-resolution data $\boldsymbol{x}_{d-1}$, $d$ day low-resolution data $\overline{\boldsymbol{x}}_d$, denoising steps $n$, and latent space denoised data $\boldsymbol{z}_{d,n}$, respectively, and the overall embedded conditions is spliced as:

$$\boldsymbol{z}_e = E_1(\boldsymbol{x}_{d-1}) \oplus E_2(\overline{\boldsymbol{x}}_d) \oplus E_3(n) \oplus E_4(\boldsymbol{z}_{d,n}) \tag{13}$$

where $\boldsymbol{z}_e$ is the embedded data, $E(\cdot)$ is the mapping function of embedding modules.

After dimension reduction using the VAE, the LDM performs denoising in the low-dimensional latent space. The dimension of the denoising noise matches the dimension of the latent space. In the embedding process, temporal convolutional layers are used to down-sample the $\boldsymbol{x}_{d-1}$, while temporal transposed convolutional layers are used to up-sample the $\overline{\boldsymbol{x}}_d$. Convolutional layers, which maintain constant dimensionality, are applied to embed the $\boldsymbol{z}_{d,n}$. For discrete denoising steps, embedding is required to map the discrete steps to align dimensions with the latent space by embedding layer and reshape layers. Finally, after adjusting the dimensions of the four embedding modules to ensure consistency, they are concatenated along the channel dimension and through the convolutional layer to generate the denoised noise $\epsilon_{\theta,n}$.

The denoising process of the diffusion model is denoted as:

$$\boldsymbol{z}_{n-1} = \frac{1}{\sqrt{\alpha_n}}(\boldsymbol{z}_n - \frac{\beta_n}{\sqrt{1-\overline{\alpha}_n}}\epsilon_{\theta,n}) + \overline{\beta}_n \epsilon \tag{14}$$

where $\alpha_n = 1 - \beta_n$, $\overline{\alpha}_n = \Pi_n \alpha_n$, $\overline{\beta}_n = \beta_n(1-\overline{\alpha}_{n-1}) \big/ 1-\overline{\alpha}_n$, and $\epsilon_{\theta,n} = \mathrm{DN}(\boldsymbol{x}_{d-1}, \overline{\boldsymbol{x}}_d, n, \boldsymbol{z}_{d,n})$.

The goal of denoising process is to expect the posterior distribution to be equal to the prior distribution. For the SRDM,

the KL divergence of $p_\theta(\boldsymbol{z}_{d,n-1} \mid \boldsymbol{x}_{d-1}, \overline{\boldsymbol{x}}_d, \boldsymbol{z}_{d,n})$ and $q(\boldsymbol{z}_{d,n-1} \mid \boldsymbol{x}_{d-1}, \overline{\boldsymbol{x}}_d, \boldsymbol{z}_{d,n})$ is used as the loss function, which is equivalent to the evidence lower bound of the likelihood function [29]. The loss function can be transformed as:

$$\begin{aligned} L_{\text{ldm}} &= KL(p_\theta(\boldsymbol{z}_{d,n-1} \mid \boldsymbol{x}_{d-1}, \overline{\boldsymbol{x}}_d, \boldsymbol{z}_{d,n}) \,||\, q(\boldsymbol{z}_{d,n-1} \mid \boldsymbol{x}_{d-1}, \overline{\boldsymbol{x}}_d, \boldsymbol{z}_{d,n})) \\ &= \mathbb{E}_{\epsilon \sim \mathcal{N}(\mathbf{0},\mathbf{1})}[\frac{\beta_n^2}{2\alpha_n(1-\overline{\alpha}_n) \,||\, \overline{\beta}_n \mathbf{I} \,||^2} \,||\, \epsilon - \epsilon_{\theta,n} \,||^2] \end{aligned} \tag{15}$$

Essentially, the denoising process is not the strict inverse of diffusion process, as the addition of random noise is unpredictable and irreversible. The denoising network, however, generates denoised noise $\epsilon_{\theta,n}$ that exhibit characteristics which can either eliminate specific features of the sample or restore them within the noise. For instance, considering temporal correlation as an example, the denoising noise will introduce a temporal correlation opposite to that of the sample in the latent space. As a result, the removal of this noise during the denoising process restores the temporal correlation, aligning it with that of the original sample. Similarly, when noise is added, the combination of the sample and the posterior noise cancels out the temporal correlation, making the samples more aligned with the prior distribution.

## IV. Numerical Results

### A. Data Source Description and Experiment Settings

The data used in this study are sourced from CMIP6 [30] and ERA5 [31] datasets. The climate data are selected from the 2025-2100 CMIP6 dataset with a temporal resolution of 1 day, while the historical reanalysis data are selected from the 2014-2023 ERA5 dataset with a temporal resolution of 1 hour. Three climate change pathways from the Canadian earth system model version 5 (CanESM5) are used: SSP126, SSP370 and SSP585. To simulate the wind and PV power, four key meteorological features are selected: east-west wind speed (U-m/s), north-south wind speed (V-m/s), solar radiation (R-W/m$^2$), and near-surface air temperature (T-C°). Table I summarizes these detailed meteorological features in the CMIP6 and ERA5 datasets.

TABLE I
Climate and Reanalysis Data Corresponding Features

| Feature | ERA5 | CMIP6 |
|---|---|---|
| U | 10m u-component of wind | U-component of wind |
| V | 10m v-component of wind | V-component of wind |
| R | Surface solar radiation downwards | Surface downwelling shortwave radiation |
| T | 2m temperature | Near-surface air temperature |

For the case study, the Ejina region in Inner Mongolia, China, is chosen as the research area. The captured geographic location is a typical undeveloped desert, gobi, and barren areas, as shown in Fig. 5. The differences in spatial resolution among the various datasets are addressed using bilinear interpolation to achieve alignment.

The generative models are implemented using the PyTorch 1.11.0 framework, and the simulations are performed on a computer equipped with the following hardware: NVIDIA RTX 3080, Intel Core i5-9400F 2.90GHz, and 64 GB RAM. The study uses the ERA5 dataset for self-supervised learning to train the SRDM model. Subsequently, the low-resolution CMIP6 data are used as input to generate high-resolution climate data. In the training process, the SRDM requires approximately 44.8 seconds per 100 epochs. To capture short-term uncertainty, the study randomly generated 100 high-resolution ensemble climate scenarios using SRDM. In the test process, SRDM takes approximately 64.9 seconds to generate one year of data for a single feature. Finally, the wind and PV power generation are converted using mechanism models, and the unit value is used to characterize the power generation capability. The source code is publicly available at: https://github.com/DREAMDXC/Super-Resolution-Recurrent-Diffusion-Model.

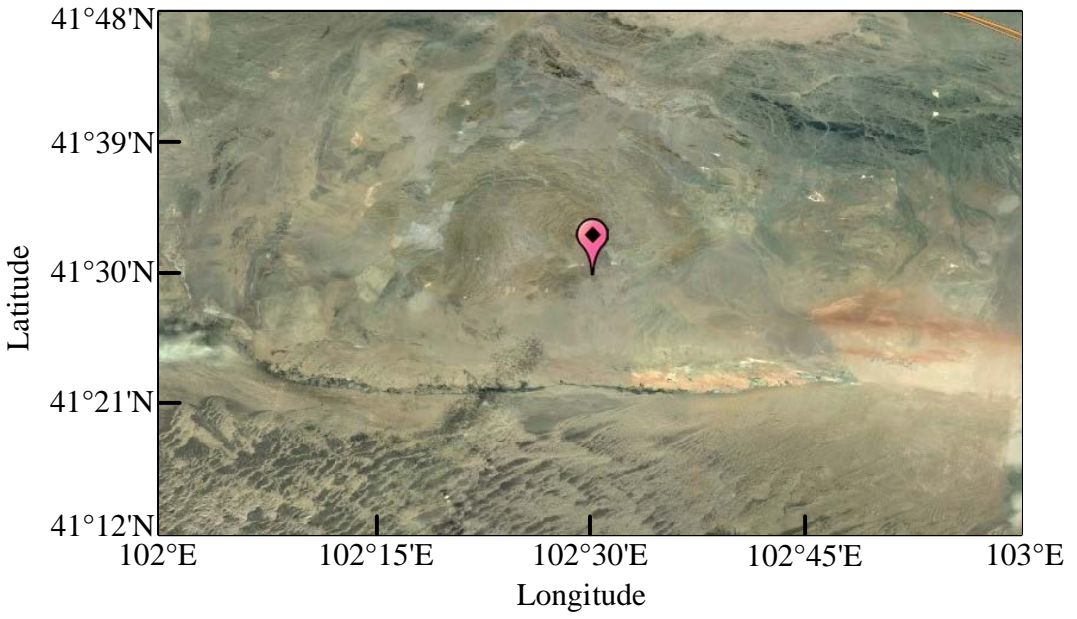

Fig. 5. Geographic map of the simulated area.

### B. Meteorology - power conversion mechanism

Wind power is converted using hub height wind speed in combination with the wind turbine power curve. The W2E-215/9.0 wind turbine model is selected, with a rate power of 9.0 MW, a cut-in wind speed of 3.0 m/s, a rated wind speed of 13.0 m/s, a cut-out wind speed of 25.0 m/s, and a hub height of 100 meters. The wind turbine power curve data are available on the open platform [32]. Since the wind speed data corresponds to the 10-meter near-surface wind speed, the wind speed at the 100-meter hub height is derived through the following conversion:

$$s_h = s_r \frac{\ln(h_c \,/\, z_0)}{\ln(h_r \,/\, z_0)} \tag{16}$$

where $s_h$ is the hub height wind speed, $s_r$ is the reference height wind speed, $h_c$ is the hub height, $h_r$ is the reference height, and $z_0$ is the surface roughness. Given the minor variation in surface roughness across the desert, gobi, and barren areas, a constant value of 0.018 is set in case study, therefore $s_h \approx 1.36 \cdot s_r$.

The power conversion of PV can be expressed as:

$$P_{\text{pv}} = f_{\text{pv}} C_{\text{pv}} \frac{G}{G_{\text{stc}}} [1 + \mu(T - T_{\text{stc}})] \tag{17}$$

where $P_{\text{pv}}$ is the PV power, $f_{\text{pv}}$ is the PV derating factor, which is related to the surface condition of the PV module and the light decay of the module, here set $f_{\text{pv}}$ =1 to simplify the conversion process, $C_{\text{pv}}$ is the rated capacity of the PV module, $G$ is the solar radiation received by the PV module, $G_{\text{stc}}$ is the standard solar radiation 1KW/m$^2$, $\mu$ is the temperature correction factor. Here the factor of LONGi Hi-MO7 module is selected, $\mu$ =-0.28%/°C [33]. $T$ is the current surface temperature of the PV module, here set $T$ equal to the air temperature, and $T_{\text{stc}}$ is the standard surface temperature 25°C.

### *C. Super Resolution Meteorological Features Results*

The trained SRDM is used to generate high-resolution ensemble climate data from the low-resolution CMIP6 data. To visualize the differences in meteorological features before and after super-resolution, the results for four meteorological features from January 2 to January 8, 2025, under the SSP126 pathway are shown in Fig. 6.

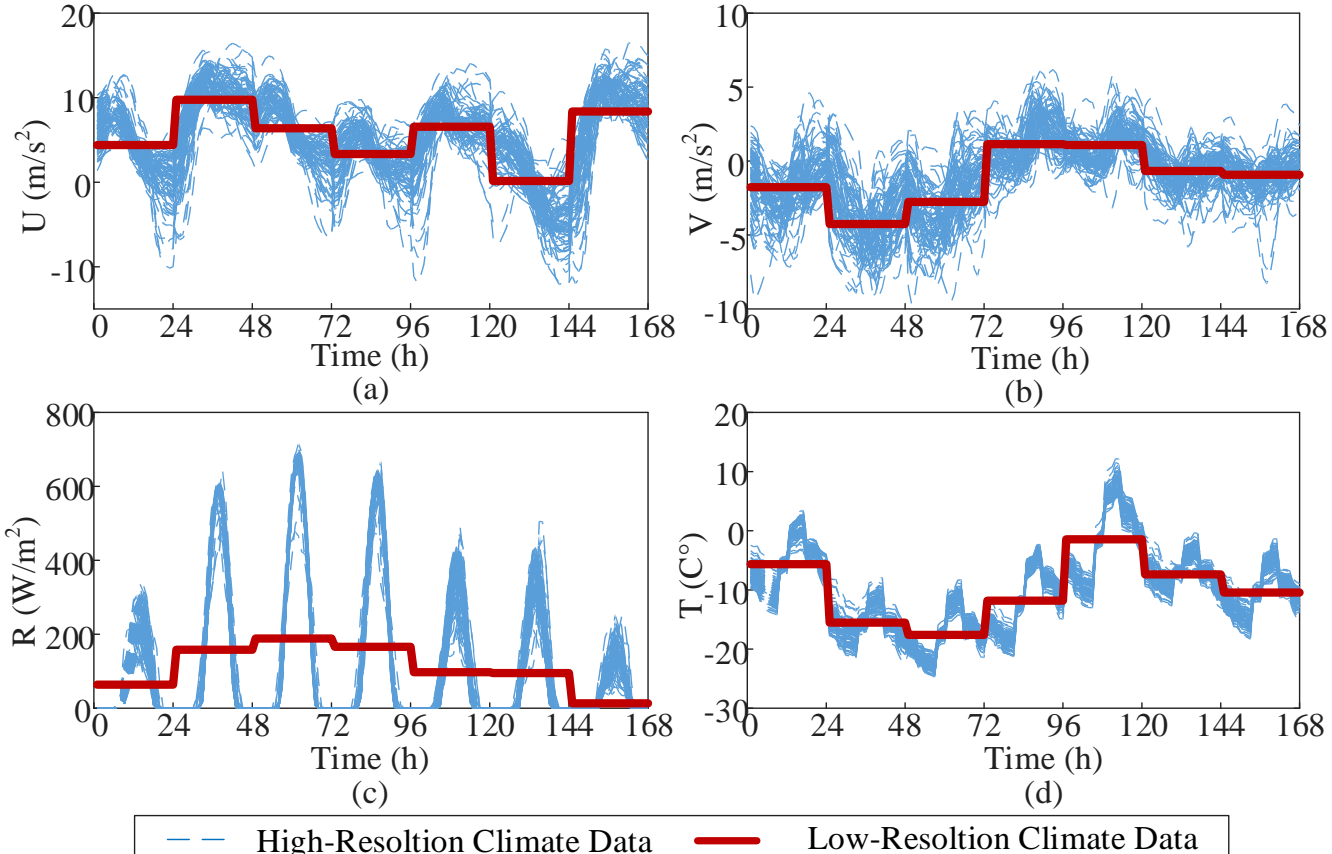

Fig. 6. Illustration of the super-resolution meteorological features.

As shown in Fig. 6, the climate data after super-resolution effectively capture the inherent fluctuations in the meteorological features. The hourly variations in radiation data during sunrise and sunset are clearly discernible, the detailed information that cannot be obtained from the low-resolution data. From the one-week high-resolution results for wind speed and temperature, it is evident that the data exhibit continuity without significant breaks between days. The high-resolution data not only provide a finer temporal scale but also maintain consistency with the overall trend of the low-resolution data. High-resolution results adjust their fluctuation patterns in alignment with changes in the low-resolution data and further reveal the uncertainties of meteorological features over short-term time scales.

To evaluate the performance of the proposed SRDM model, three widely used super-resolution methods were selected as benchmarks: ESRGAN [34], VAESR [35], and SRDiff [36]. The consistency of the super-resolution results with the low-resolution data was assessed using the mean absolute error (MAE) between the daily meteorological features. The results are shown in Table II.

TABLE II
THE DEVIATION RESULTS OF DAILY METEOROLOGICAL FEATURES

| Model | U | V | R | T |
|---|---|---|---|---|
| ESRGAN | 1.254 | 0.824 | 6.071 | 0.622 |
| VAESR | 0.875 | 0.542 | 5.588 | 0.325 |
| SRDiff | 0.391 | 0.286 | 5.085 | 0.182 |
| SRDM | 0.218 | 0.191 | 4.043 | 0.134 |

Table II shows that the proposed SRDM model exhibits the lowest MAE values when compared to the other methods for the four meteorological features. Most existing super-resolution models are primarily focused on image applications, with less attention given to long-term time series data. These models often struggle to maintain continuity when generating super-resolution data across daily data. In contrast, SRDM not only reconstructs low-resolution climate data more accurately as boundary conditions for super-resolution but also performs significantly better in maintaining data coherence across different days.

Furthermore, the generated super-resolution climate data were used to analyze the changes in the annual mean trends of meteorological features. During the analysis, east-west and north-south wind speeds were considered after taking their absolute values. The Theil-sen estimator was used to estimate the trend changes of meteorological features under three pathways. Fig. 7 illustrates the trends of these meteorological features, with boxplots corresponding to 100 climate scenarios and the fitted line for the period from 2025 to 2100. Table III provides the corresponding quantitative data.

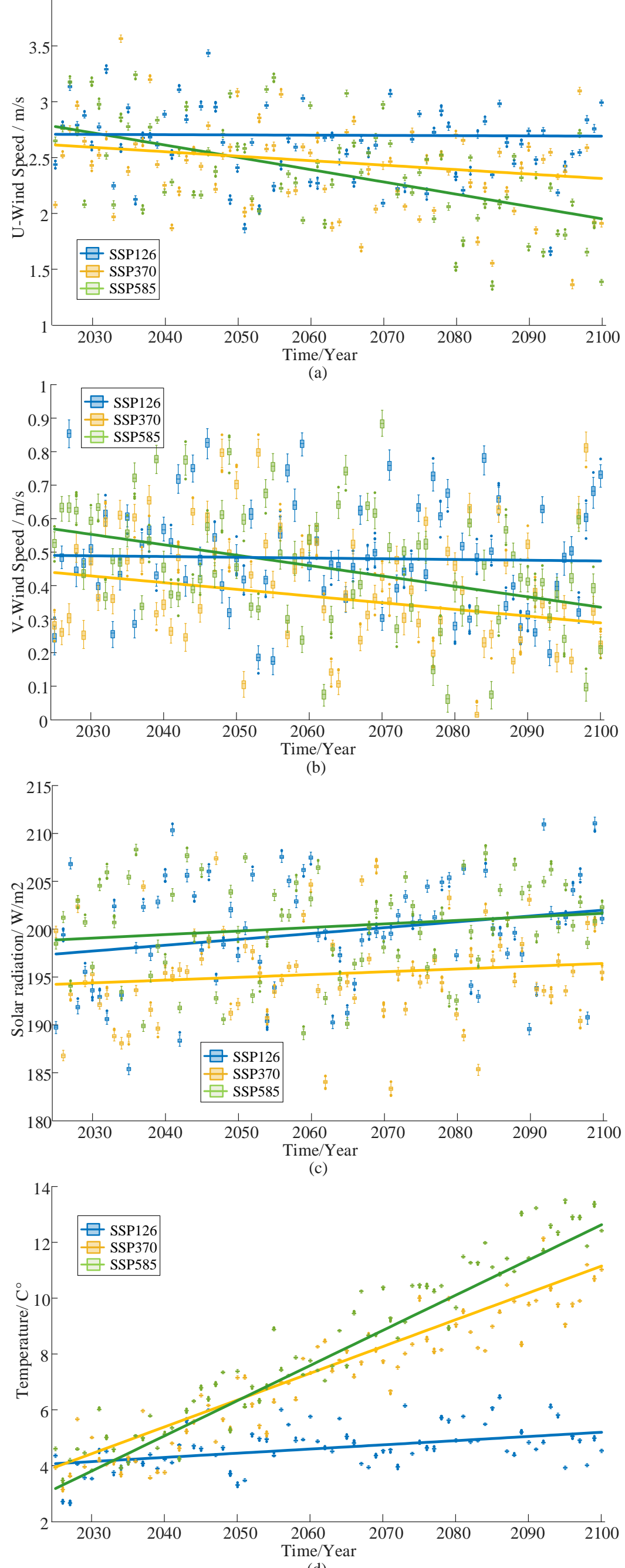

Fig.7 Meteorological feature trends in 2025-2100.

Fig. 7 shows that east-west wind speed and north-south wind speed tend to stabilize under SSP126 pathway, whereas exhibit a decreasing trend under SSP370 and SSP585 pathway. Moreover, it can be observed that the wind energy resources in the Ejina region are more concentrated in the east-west direction. Global warming in SSP585 is expected to lead to a decrease in wind speed in Inner Mongolia, with a reduction in east-west wind speed of -0.011 m/(s·year) and a mean reduction in north-south wind speed of -0.003 m/(s·year). Although the decrease in mean wind speed is not substantial, wind power generation is proportional to the cube of wind speed, so even a slight decrease can result in significant changes in wind power generation.

In contrast, solar radiation shows no significant trend under three pathways. The temperature factor exhibits the most pronounced trend, with SSP126 indicating a global temperature rise of up to 2°C by the end of the century, while SSP585 shows a strong temperature increase in the Inner Mongolia region, with an annual rise of 0.126°C, leading to a temperature increase of more than 4°C by 2060.

TABLE III

TREND ESTIMATION RESULTS FOR METEOROLOGICAL FEATURES

| Feature | SSP | Trend | Slope | Intercept |
|---|---|---|---|---|
| U | 126 | no trend | -0.002 | 2.709 |
| | 370 | decreasing | -0.004 | 2.617 |
| | 585 | decreasing | -0.011 | 2.789 |
| V | 126 | no trend | 0.000 | 0.490 |
| | 370 | decreasing | -0.002 | 0.441 |
| | 585 | decreasing | -0.003 | 0.571 |
| R | 126 | no trend | 0.061 | 197.343 |
| | 370 | no trend | 0.029 | 194.210 |
| | 585 | no trend | 0.037 | 198.840 |
| T | 126 | increasing | 0.015 | 4.057 |
| | 370 | increasing | 0.096 | 3.856 |
| | 585 | increasing | 0.126 | 3.055 |

### *D. Power Generation Conversion Results*

After obtaining the high-resolution climate scenario dataset, the hourly power generation curves for the Ejina region from 2025 to 2100 were calculated using the power conversion mechanism model. Fig. 8 presents the results of wind and PV power generation from January 2 to January 8, 2025, under the SSP126 pathway. The figure clearly shows the representation of short-term uncertainty in renewable power generation.

Additionally, consistent with the meteorological feature analysis, Terrell-Son estimator was used to estimate the trend changes of annual utilization hours (AUH) in wind and PV power generation from 2025 to 2100, as shown in Fig. 9. Table IV provides the corresponding quantitative data, where the interval width is used to assess the uncertainty in power generation.

As shown in Fig. 9(a), there is no significant trend change in wind power generation under the SSP126 and SSP370 pathways, whereas a substantial decline is observed under the SSP585 pathway. Specifically, under global warming due to uncontrolled emissions in SSP585, the annual wind power generation hours decrease by approximately 2.819 hours/year. This implies a cumulative decrease of about 100 hours in annual wind power utilization in the Ejina region by 2060. This trend aligns with the observation that wind power amplifies fluctuations in wind speed. Moreover, Table IV shows that for a given climate trend, the fluctuation interval width in annual wind power generation hours exceeds 300 hours, driven by short-term uncertainties. This results in an uncertainty in power generation of more than 7%, that is not captured by low-resolution climate simulations.

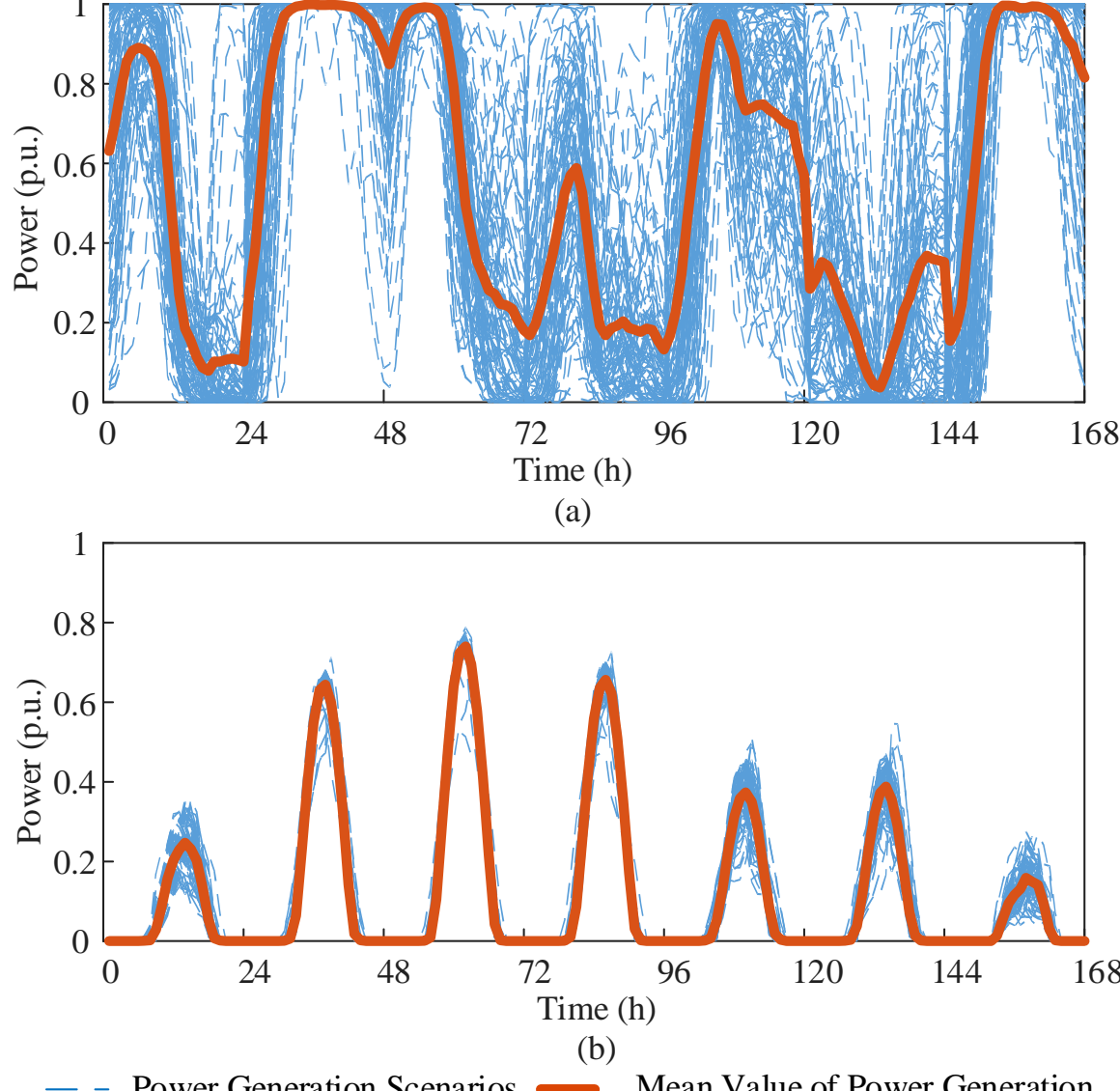

Fig.8 Illustration of the renewable energy power generation.

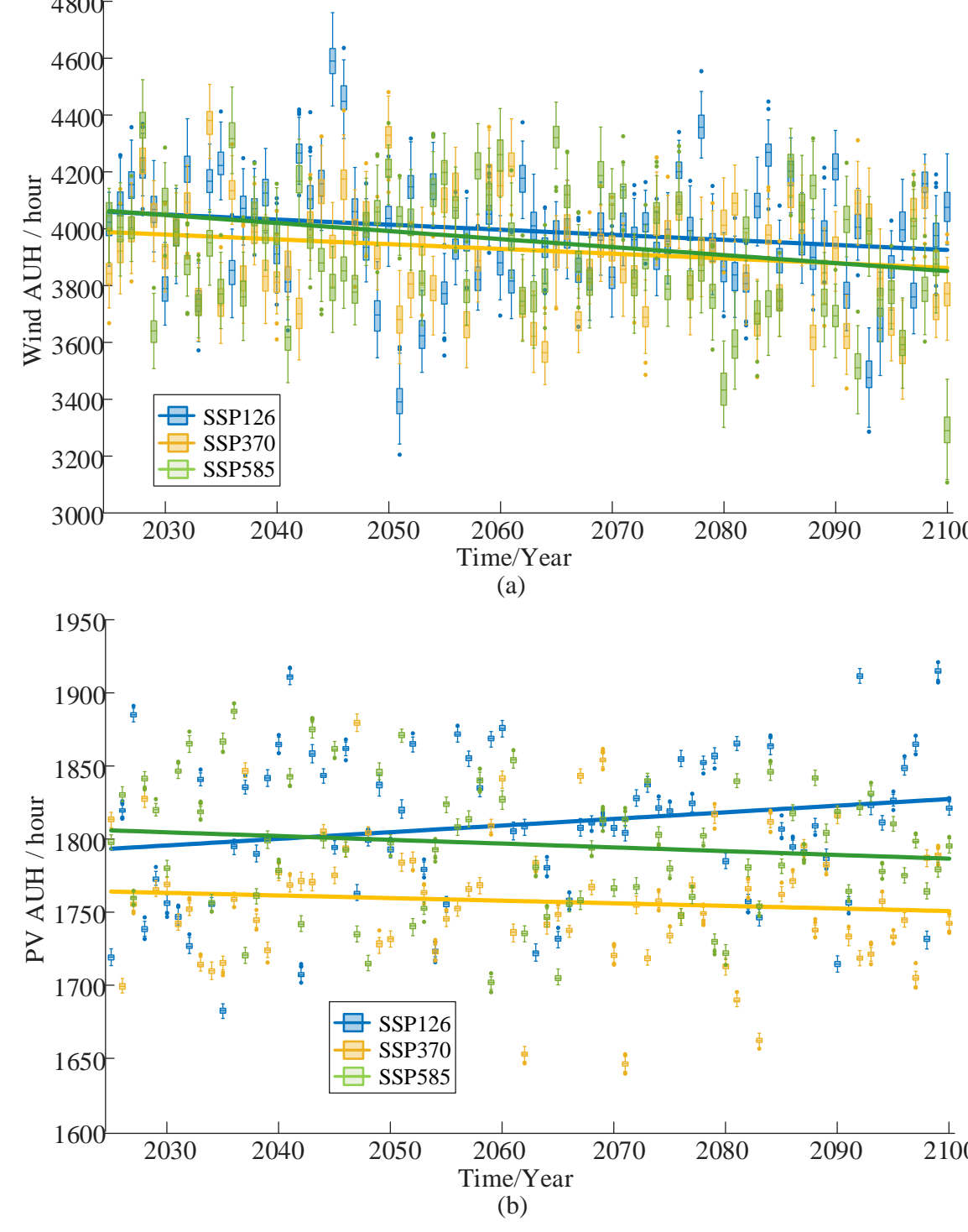

Fig.9 Power generation trends in 2025-2100.

Fig. 9(b) shows that the trend for PV power generation remains insignificant under both the three pathways. However, under SSP 126 pathway, where global warming is effectively controlled, the AUH of PV power increase by +0.453 hours/year, driven by a slight upward trend in solar radiation. In contrast, under SSP370 and SSP 585 pathways, where global warming is more severe, significant rise in temperature may reduce the efficiency of PV modules. Extreme high temperatures under SSP585 pathways lead to a decrease in PV power AUH of -0.258 hours/year. Although the change in PV power generation is relatively minor and the fluctuation interval width in AUH due to short-term uncertainty being approximately 11 hours (accounting for no more than 1%), the divergence in PV power generation trends between the climate pathways remains considerable. By the end of the century, the difference in PV AUH between these pathways reaches approximately 60 hours. It is important to note that the current simulations do not account for the potential impact of rapidly rising temperatures on the operational reliability of PV modules. This effect may be exacerbated at the gobi, dust, and desertification energy base.

Therefore, from the perspective of long-term power system planning, it is crucial to consider the impact of future climate change on the power generation of energy bases, as this will directly affect the power system's ability to balance electricity supply and demand.

TABLE IV
TREND ESTIMATION RESULTS FOR POWER GENERATION

| Feature | SSP | Trend | Slope | Intercept | Interval width |
|---|---|---|---|---|---|
| Wind | 126 | no trend | -1.783 | 4060.738 | 322.623 |
| | 370 | no trend | -1.679 | 3989.529 | 324.379 |
| | 585 | decreasing | -2.819 | 4065.224 | 317.541 |
| PV | 126 | no trend | +0.453 | 1792.896 | 11.972 |
| | 370 | no trend | -0.178 | 1764.251 | 11.819 |
| | 585 | no trend | -0.258 | 1806.171 | 11.593 |

### *E. Various Resolution Conversion Results*

To compare the effects of power conversion in different resolutions climate data, we evaluate the difference in AUH using both low- resolution and high-resolution climate data. Fig. 10 shows the deviation values for each year in the form of a scatter plot. The values are low resolution conversion results minus high resolution conversion results.

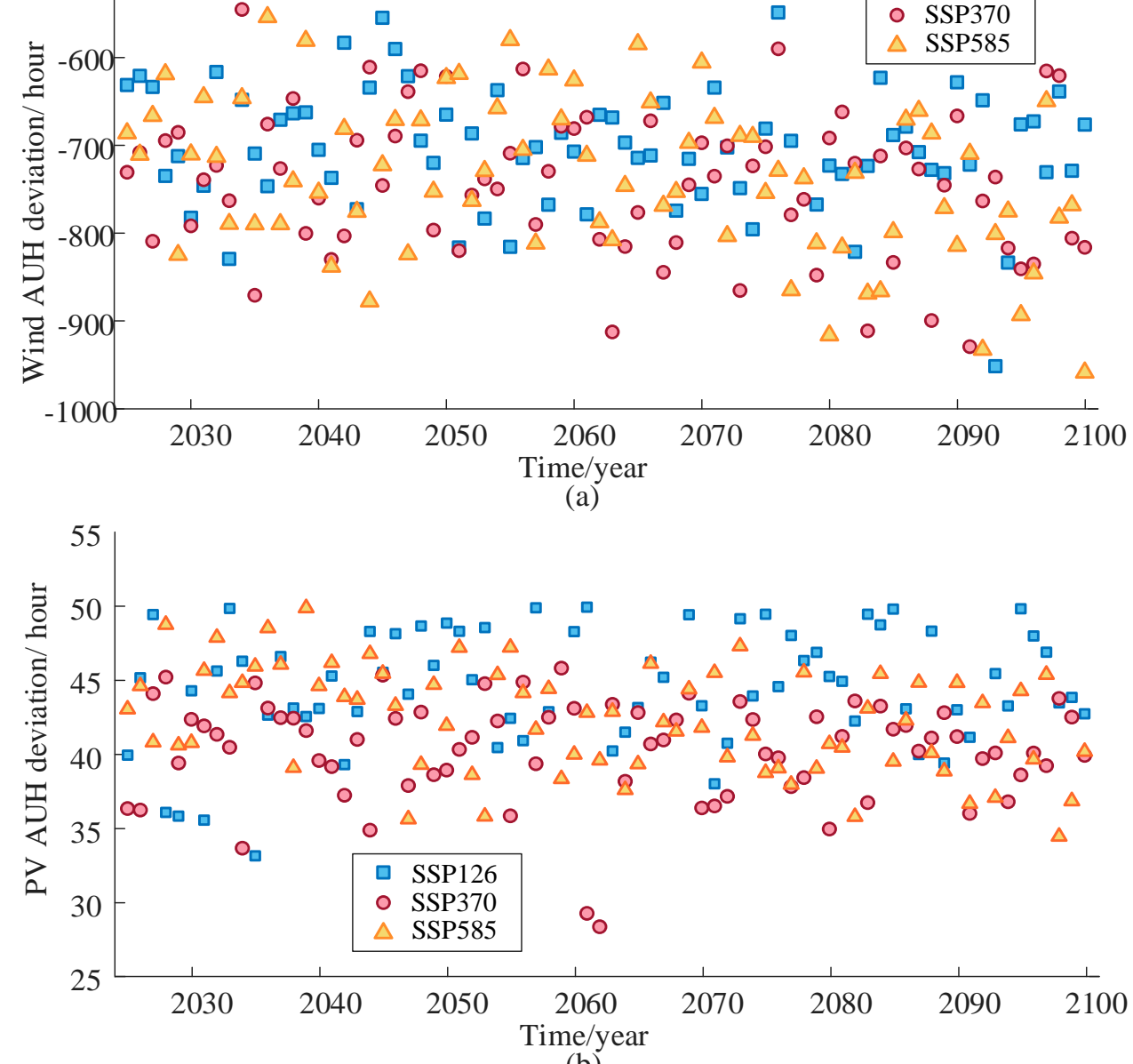

Fig.10 Illustration of various resolution conversion results.

As shown in Fig. 10(a), power conversion using the CMIP6 raw daily wind speed data results in a significant underestimation of the wind power AUH, with deviations ranging from -12.02% to -29.15%. Based on the wind turbine power curve, wind power output exhibits a nonlinear relationship with wind speed. When the wind speed is below the cut-in speed, the power output is zero. Therefore, using average wind speed data for power conversion seriously underestimates the effective wind speed available for wind turbine power generation.

As shown in Fig. 10(b), power conversion using the CMIP6 raw daily radiation and temperature data leads to an overestimation of the PV power AUH, with deviations ranging from +1.69% to +2.88%. In contrast to wind power, the AUH of PV exhibit a smaller deviation, primarily due to the more linear correlation between the PV power generation and solar radiation. However, at night, when temperatures are lower and no radiation, using average temperatures for power conversion underestimates the temperature values during actual daylight power generation hours. This underestimation leads to an increase in the PV power AUH, as the efficiency of PV modules decreases at higher temperatures.

Based on the above analysis, it can be concluded that directly using low resolution climate data for power conversion is not recommended for either wind or PV power generation, as it introduces additional biases, particularly in wind power. Therefore, super-resolution modeling of low-resolution climate data is necessary to simulate long-term generation in renewable energy.

### *F. Effect of module parameters on AUH*

In recent years, wind turbine manufacturers have focused on reducing cut-in wind speeds, enabling turbines to operate under lower wind speed. Similarly, PV module manufacturers have focused on lowering the temperature coefficient factor, enhancing the efficiency of PV modules at higher temperatures. Looking ahead, technological advancements in these module parameters are expected to interact with climate change, creating compounding effects on renewable energy generation. This study examines the impact of these parameter changes on AUH, specifically the cut-in wind speed of wind turbines and the temperature coefficient factor of PV modules. Fig. 11 illustrates the effects of varying these parameters under the SSP 585 pathway. The cut-in wind speed is analyzed across a range of 2.6 m/s to 3.4 m/s, while the temperature coefficient factor is analyzed across a range of -0.24%/°C to -0.32%/°C.

As shown in Fig. 11, lowering the cut-in wind speed of wind turbines increases the AUH of wind power by enabling turbines to operate at lower wind speeds. Since global warming reduces the availability of high wind speed resources, optimizing wind turbine performance for lower wind speeds can effectively mitigate the impact of declining wind resources. At a cut-in wind speed of 2.6 m/s, the slope in AUH is mitigated by 0.191 hours/year compared to a cut-in wind speed of 3.0 m/s. In the case of PV power generation, a lower temperature coefficient factor helps minimize the adverse effects of elevated temperatures on PV efficiency. At a coefficient factor of

-0.24%/°C, the slope in AUH is mitigated by 0.08 hours/year compared to a coefficient factor of -0.28%/°C.

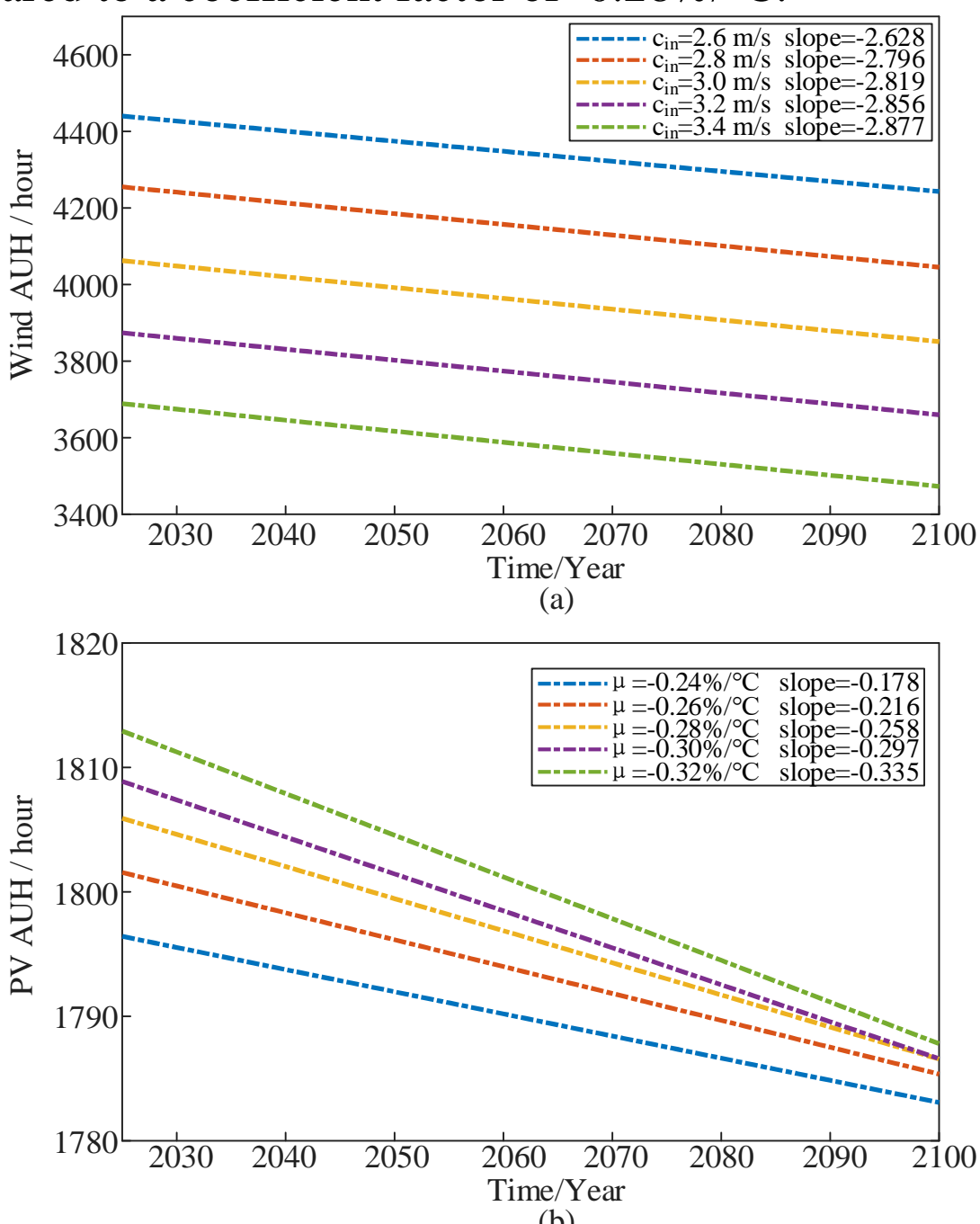

Fig.11 Illustration of various module parameters AUH results.

## V. Conclusions

In this paper, we propose the SRDM which enhances the resolution of daily climate data to an hourly resolution. The SRDM model not only improves the temporal resolution of climate data but also effectively characterizes the uncertainty factors in short-term meteorological features. The high-resolution climate data is more suitable for evaluating renewable energy generation. Compared with existing models, SRDM uses a perceptual VAE that maps a high-dimensional space to a low-dimensional latent space, thereby decoupling temporal correlations and reducing the training complexity of the diffusion model. Furthermore, by introducing a recurrent diffusion mechanism, SRDM is able to maintain continuity among samples in time-series super-resolution tasks over long time scales, thus avoiding discontinuities in the generated data.

In the case study, the SRDM was applied to enhance the temporal resolution of climate data for the Ejina region in Inner Mongolia, China, followed by power generation conversion using a mechanistic model. The results demonstrate that the proposed approach effectively simulates hourly meteorological features and converts them into power generation data, making it more suitable for power system operation simulations. The findings reveal that directly using low-resolution climate data for power generation conversion introduces significant biases. Additionally, enhancing the performance parameters of renewable energy modules was shown to mitigate the impact of climate change on AUH.

Future research will focus on enhancing modeling accuracy in two critical areas. First, improving the representation of transient processes in power generation equipment is essential. Current mechanistic models primarily address steady-state conditions and often neglect transient dynamics, which may lead to an overestimation of AUH. Second, renewable energy generation is affected by numerous factors beyond meteorological variables, including equipment types, installation methods, and operational modes. Consequently, the development of mechanistic models that comprehensively incorporate these factors is crucial for achieving more precise assessments of renewable energy output.

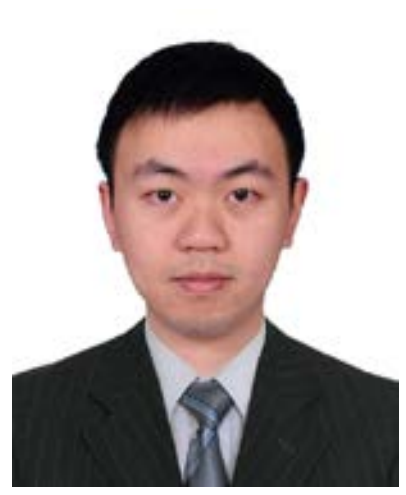

**Xiaochong Dong** received the B.S. degree from China Three Gorges University, Yichang, China, in 2017, the M.S. degree in North China Electric Power University, Beijing, China, in 2020, and the Ph.D. degree in North China Electric Power University, Beijing, China, in 2024, both in electrical engineering. He is currently a postdoctoral researcher at Tsinghua University, Beijing, China. His research interests include artificial intelligence and optimization of the power system.

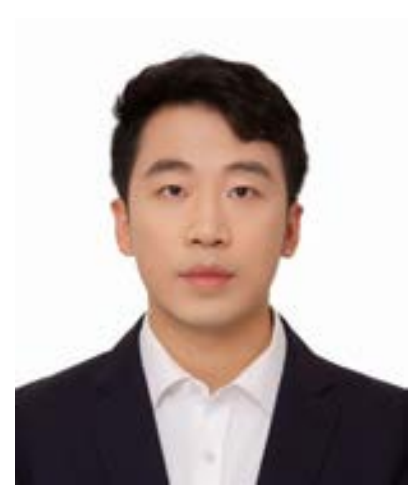

**Jun Dan** is a Ph.D. candidate at the College of Information Science & Electronic Engineering, Zhejiang University, China. He received the B.Eng. degree in Electronic Information Engineering from Chongqing University, China, in 2020. He serves as a reviewer for prestigious computer vision conferences and journals, including CVPR, ECCV, ICCV, TIP, ICLR, NeurIPS, and ACM MM. His research interests include transfer learning, face recognition, and LMMs.

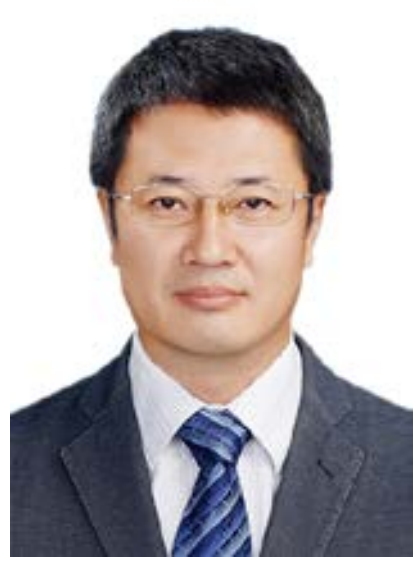

**Yingyun Sun** received the B.S. degree from Shanghai Jiao Tong University, Shanghai, China, in 1996, M.S. and Ph.D. degrees from Tsinghua University, Beijing, China, in 2004 and 2007, respectively, all in electrical engineering. He is currently a Professor at the School of Electrical Engineering at North China Electric Power University, Beijing, China. His research interests include artificial intelligence, renewable energy, and power system analysis.

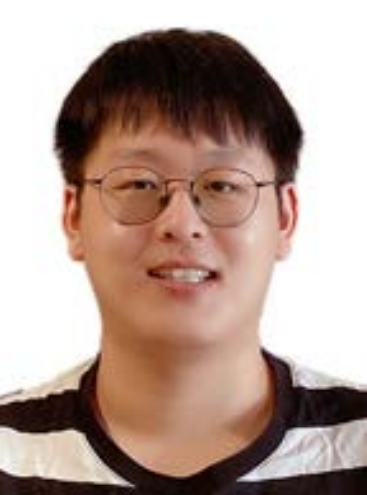

**Yang Liu** is a Ph.D. candidate at the Department of Engineering, King's College London. He received his M.Sc. degree in automation from North China Electric Power University, China in 2020. He is a reviewer in prestigious computer vision conferences including CVPR, ICCV, ICML, ICLR, and NeurIPS. His main research interest is deep face perception, robotics, computer vision and VLMs. He has published over 20 articles in prestigious international journals and conferences.

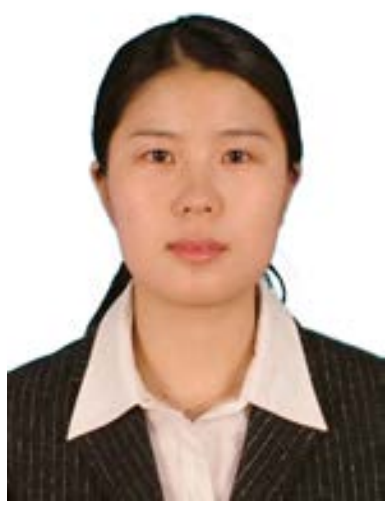

**Xuemin Zhang** received the B.Sc. and Ph.D. degrees in electrical engineering from Tsinghua University, Beijing, China, in 2001 and 2006, respectively. She is currently an Associate Professor with Tsinghua University. Her research interests include power system stability and control, renewable energy forecast, and power system complexity.

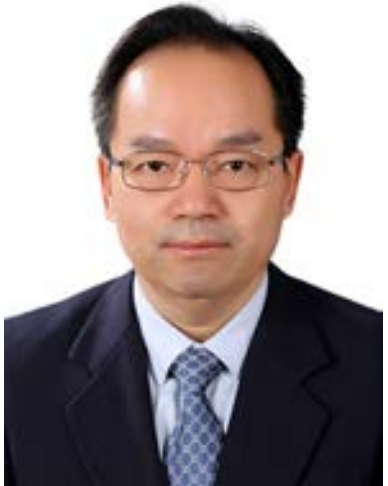

**Shengwei Mei** received the B.Sc. degree in mathematics from Xinjiang University, Urumqi, China, in 1984, the M.Sc. degree in operations research from Tsinghua University, Beijing, China, in 1989, and the Ph.D. degree in automatic control from the Chinese Academy of Sciences, Beijing, in 1996. He is currently a Professor with Tsinghua University. He is also the Director of the New Energy Industry Research Center, Qinghai University, Xining, China. His research interests include power system analysis and control, robust control of power systems, comprehensive utilization of new energy, and disaster prevention of large power grids.